\documentclass{article}

\usepackage{arxiv}
\usepackage{multirow}
\usepackage{subfig}
\usepackage{graphicx}
\usepackage{amsmath}
\usepackage{amssymb} 
\usepackage{pifont}
\usepackage[utf8]{inputenc} % allow utf-8 input
\usepackage[T1]{fontenc}    % use 8-bit T1 fonts
\usepackage{hyperref}       % hyperlinks
\usepackage{url}            % simple URL typesetting
\usepackage{booktabs}       % professional-quality tables
\usepackage{amsfonts}       % blackboard math symbols
\usepackage{nicefrac}       % compact symbols for 1/2, etc.
\usepackage{microtype}      % microtypography
\usepackage{cleveref}       % smart cross-referencing
\usepackage{lipsum}         % Can be removed after putting your text content
\usepackage{graphicx}
\usepackage{natbib}
\usepackage{doi}

\title{Adaptive Image Inpainting}

\date{}

\author{ \hspace{1mm} Maitreya Suin
	\And \hspace{1mm} Kuldeep Purohit \And
	\hspace{1mm}A. N. Rajagopalan
}

\renewcommand{\headeright}{}
\renewcommand{\undertitle}{}
\begin{document}
\maketitle

\begin{abstract}
	Image inpainting methods have shown significant improvements by using deep neural networks recently. However, many of these techniques often create distorted structures or blurry textures inconsistent with surrounding areas. The problem is rooted in the encoder layers' ineffectiveness in building a complete and faithful embedding of the missing regions. To address this problem, two-stage approaches deploy two separate networks for a coarse and fine estimate of the inpainted image. Some approaches utilize handcrafted features like edges or contours to guide the reconstruction process. These methods suffer from huge computational overheads owing to multiple generator networks, limited ability of handcrafted features, and sub-optimal utilization of the information present in the ground truth. Motivated by these observations, we propose a distillation based approach for inpainting, where we provide direct feature level supervision for the encoder layers in an adaptive manner. We deploy cross and self distillation techniques and discuss the need for a dedicated completion-block in encoder to achieve the distillation target. We conduct extensive evaluations on multiple datasets to validate our method.
\end{abstract}

% keywords can be removed
%\keywords{First keyword \and Second keyword \and More}

\section{Introduction}
Image inpainting intends at filling damaged or substituting undesired areas of images with plausible and fine-detailed contents. It has a broad range of applications in fields of restoring damaged photographs, retouching pictures, etc. Early conventional works usually make use of low-level features (e.g., color and texture descriptors) hand-crafted from the incomplete input image and resort to priors (e.g., smoothness and image statistics) or auxiliary data (e.g., external image databases). They either propagate low-level features from surroundings to the missing regions following a diffusive process \cite{bertalmio2000image,roth2005fields} or fill holes by searching and fusing similar patches from the same image or external image databases \cite{hays2007scene,pritch2009shift}. While these methods are promising for some cases like background inpainting tasks, it is restricted by the available image statistics. It cannot capture the high-level semantics or global structure of the image. Another notable drawback of traditional diffusion-based and patch-based methods is that they assume missing patches can be found somewhere in the background regions. So, for situations involving intricate structures like faces \cite{yu2018generative} and complex inpainting regions, they cannot produce novel image contents.

\par In recent years, the deep learning based methods have been reported to surmount the limitations by utilizing the large volume of training images. The encoder-decoder architecture is prevalent in existing deep inpainting methods \cite{li2017generative,yeh1607semantic}. However, direct utilization of the end-to-end training and prediction processes generates limited results due to the severe ill-posedness of the task. The hole regions are entirely empty, and without sufficient guidance, an encoder-decoder cannot reconstruct the missing content adequately.

\par To alleviate this difficulty, two-stage methods have been proposed to do content formation and texture refinement separately in a step-by-step manner. These two-stage methods \cite{yu2018generative,yu2019free,liu2019coherent} typically produce an intermediate coarse image with recovered structures in the first stage and send it to the second stage for texture generation. Recently, \cite{nazeri2019edgeconnect} propose to utilize explicit image structure knowledge for inpainting. They develop a two-stage model that comprises an edge generator followed by an image generator. The edge generator is trained to hallucinate the possible edge forms of the masked areas, which act as the precondition for the generator network. For scenes containing more salient objects, \cite{xiong2019foreground} proposed a contour generator instead of the edge generator.

\par We build our approach based on two key insights: a) The ground-truth image contains ample information, and it can be exploited to directly guide the encoder. b) Generally, deeper layers of an inpainting network contain more complete information of the missing regions. Intuitively, we follow the idea that learning can be enhanced by guiding the learner through intermediate tasks. Training deep architectures is easier when some hints are given about the function that the intermediate levels should compute. The quality of learning significantly improves when provided with intermediate supervisions than to expect the network to learn it automatically only from the final supervision at the end. In this work, we resort to the use of knowledge distillation technique to provide intermediate supervisions for the encoder layers.

\par We show two entangled distillation techniques: cross (inter-network) and self (intra-network) distillation and how they operate together to achieve the goal. For cross distillation, we transfer the knowledge from an auxiliary network that contains full information about the missing regions' features. An off-the-shelf under-complete autoencoder is trained to reconstruct the GT image. The intuition is: in a supervised setting, abundant high-quality information is available at training time, which can be extracted from the GT image itself. We assume that the features extracted from the GT image, containing uncorrupted information about the missing regions, can be used to directly supervise the inpainting encoder. We break the encoder in small stages and use the supervision at each stage. Next, to mine for more task-specific knowledge, we use self-distillation within different levels of the inpainting network. Generally, as we go deeper, the hole regions get filled gradually and the deeper features more closely resembles the actual missing information. Taking the motivation from this observation, we propose the self distillation framework, that distills knowledge within the inpainting network itself and the knowledge in the deeper portion of the networks is squeezed into the shallow ones. The main difference from the conventional knowledge distillation is that the teacher is not a static model, but dynamically evolves as training proceeds. Cross distillation will improve the encoder layer's performance, and such improvement will further flow from the deeper to shallower layers by self distillation. 

\par This distillation based approach provides a direct target for the encoders and using it as-it-is gives a boost to the inpainting network's performance. But, we argue that, this still might not be the optimal approach. The reconstruction process heavily depend on the regions surrounding a hole and to get a better understanding of the surrounding regions, encoder is also expected to extract deep meaningful features from the unaffected areas. Now, it poses additional challenge on the same encoder to simultaneously master two tasks: extract features from the uncovered regions, and use that information to fill corrupted regions. To address this issue, we discuss an elegant architectural modification and propose dedicated 'filler-block' in the encoder, whose sole purpose is to use the already available information to fill the holes. We use a parallel lightweight residual block at each level, which takes the encoded feature till that point, deploys adaptive convolution layer with variable weight and offset and solely updates the hole regions. This design makes it much easier for the encoder to reach the target provided by the distillation techniques. 

\par Our main contributions are as follows:
\begin{itemize}
	\item We propose a distillation based approach for inpainting, which utilizes deep feature level guidance for the encoder layers, resulting in the generation of better embedding of the hole regions.
	\item To achieve the distillation target more effectively, we propose the use of dedicated filler-blocks in the encoder which solely focus on updating the hole regions.
\end{itemize}

\section{Related Works}
\textbf{Image Inpainting}: Deep learning-based image inpainting approaches \cite{li2017generative,pathak2016context} are generally based on generative adversarial networks (GANs) to generate the pixels of a missing region. \cite{yan2018shift} and \cite{yu2018generative} devised feature shift and contextual attention operations, respectively, to allow the model to borrow feature patches from distant areas of the image. Some traditional approaches can also be found in the literature, such as \cite{bhavsar2012towards,kulkarni2013depth,sahay2013lost,sahay2015geometric}.  \cite{liu2019coherent} used a coherent semantic attention layer to ensure semantic relevance between swapped features. \cite{xiong2019foreground,nazeri2019edgeconnect} filled images with contour/edge completion and image completion in a step-wise manner. \cite{ren2019structureflow} first predicted smooth structure and used that for the final stage. \cite{zhang2018semantic} used cascaded generators to progressively fill in the image. These approaches attempted to solve inpainting tasks by adding structural constraints, but they still suffer from the limitation of handcrafted guidance, huge load of two generators and lack of local semantic consistency.
\newline \textbf{Distillation}: Distillation technique is mainly used in image classification \cite{hinton2015distilling}, image segmentation \cite{liu2019structured}, etc. \cite{urban2016deep} proposed ensemble of teacher, \cite{mirzadeh2019improved} showed cascaded distillation technique. Few recent works \cite{bagherinezhad2018label,furlanello2018born} have shown that self-distillation can improve the student over the teacher. \cite{zagoruyko2016paying}) force the student to mimic the activation map of the teacher (norm across channel in each spatial location). In this work, we explore knowledge distillation for supervising an inpainting network.

\section{Method}
We use an standard encoder-decoder architecture as the backbone of our inpainting network. At every encoder level, we use the proposed completion block to focus on filling the holes. The attention module together with the adaptive global-local consistent structure is used in the decoder. We follow similar encoder decoder architecture for the autoencoder (AE) branch. The AE is trained with the objective of reconstructing the ground-truth image. Due to its under completeness, it learns to extract relevant features from the image and utilize it.

\begin{figure*}
	\centering
	\includegraphics[width = 0.88\textwidth,height = 0.3\textwidth]{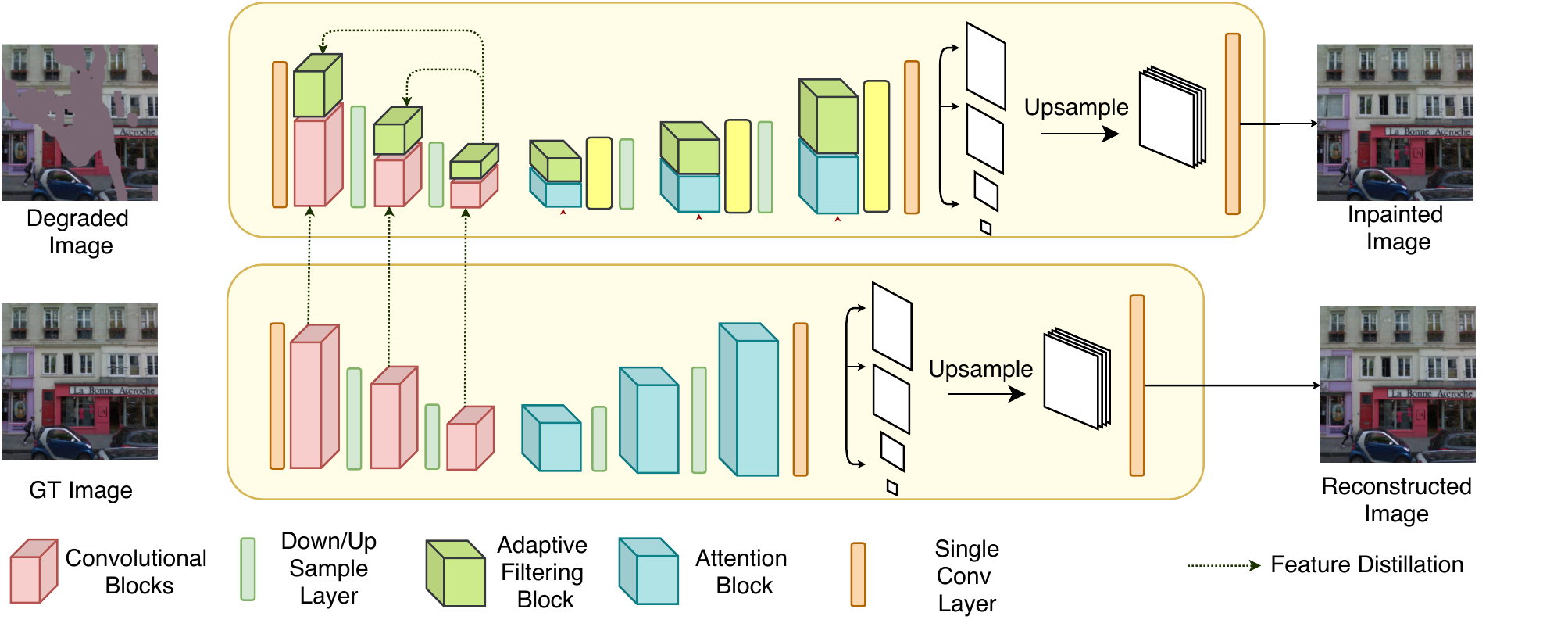}
	\caption{An overview of the network architecture.}
	\label{fig:my_label}
\end{figure*}

\subsection{\textbf{Knowledge-transfer}}
\label{sec:knowledge_transfer}
We want the inpainting encoder to mimic the performance of the AE encoder for the hole regions. This is a form of regularization, and thus, the pair hint-guided layer has to be chosen such that the inpainting network is not over-regularized and stuck in a non-optimal minima. Note that how to break down AE and inpainting network is worth a study by itself, but is beyond the scope of this work. Using a similar encoder-decoder structure for both the modules, we use the down-sampling layers of the network as the breakpoints. We observed this option to work fairly well in practice. However, considering the difference in tasks, this might not be the best design choice. Every feature channel of the AE encoder might not be equally beneficial. To tackle this problem, we use meta-networks \cite{jang2019learning} to decide which feature maps (channels) of the AE model are useful and relevant for the inpainting task. If there are $L$ encoder levels/breakpoints, then for each of the $L$ pairs, we introduce transfer importance predictor, which enforces different penalties for each channel according to their utility for the target task. For any particular pair, we have
\vspace{-2mm}
\begin{equation}
	R(\theta,x,\rho^{l})^{l}_{cross} = \sum_{c\in C} \rho^{l}_c(x_l^*) ||(x_l - x_l^*)_c \odot M ||^2_2
\end{equation}
\vspace{-2mm}
where $l \in \{ 1,L\}$ and $\rho^{l}_c : \mathbb{R}^{C \times H \times W} \rightarrow \mathbb{R}^{C}$ is the non-negative weight of channel $c$ with $\sum_{c\in C} \rho^{l}_c = 1$. For any tensor $z$, the term $z_c$ denotes the $c^{th}$ slice of the tensor. 
\newline Self distillation technique has been previously explored in image classification \cite{yun2020regularizing}, language modeling \cite{hahn2019self}, etc. The utility of self-distillation in our approach is based on the key observation that generally the deeper layers of an inpainting network contain more complete information for the hole regions.
We use L2 loss between features maps of different depths. This will force the shallow layers to mimic the behavior of the deeper layer and in turn generate better content for the hole regions. Note that, the feature maps in different depths have different sizes, so we add extra layers to align them while training. Given $x_l$ to be the encoder feature from the $l^{th}$ level, the self distillation loss can be expressed as
\vspace{-2mm}
\begin{equation}
	R(\theta,x,\phi^{l})^{l}_{self} = \sum_{c\in C} \phi^{l}_c(x_{d}) ||(f_l(x_l) - x_{l+1})_c \odot M ||^2_2
	\vspace{-1mm}
\end{equation}
where similarly $l \in \{ 1,L-1\}$ and $\phi^{l}_c : \mathbb{R}^{C \times H \times W} \rightarrow \mathbb{R}^{C}$ is the non-negative weight of channel c with $\sum_{c\in C} \phi^{l}_c = 1$, $f_m$ is the convolutional operation to make $x_l$ of the same dimension as $x_{l+1}$.
\newline \textbf{Filler-block in Encoder:}
\label{sec:adaptive_conv}
Distillation provides a direct target for the encoder, and using it results in immediate boost in the performance. But, it still remains a challenge for the encoder to achieve the goal. The encoder layers should excel at extracting meaningful deep features, especially for the unmasked areas, as those features are the only source of accessible information while filling out the holes. Better features suggest a better understanding of the whole scene content, and these even play a notable role in affinity finding attention module in the decoder. Intuitively, the two tasks: feature extraction and filling the holes are not very close. Relying on the same encoder layers to effectively meet the additional distillation objective will be sub-optimal in exploiting that feature supervision. Mask-based convolution is proposed in \cite{liu2018image,yu2019free} but they still utilize same layers for both type of regions. We propose dedicated completion-blocks at each level of the encoder which solely update the hole region. It not only reduces the load on the main encoder layers, which can focus on feature extraction, but it also makes the completion blocks to specialize in filling holes. Ultimately, this design choice makes it much easier to reach the distillation target. If $E_{m-1}$ is the feature before the $m^{th}$ level of the encoder, output of $m^{th}$ level, $E_{m}$ can be expressed as
\begin{equation}
	E_{m} = f_{feat}(E_{m-1}) + f_{fill}(E_{m-1}) \odot M
\end{equation}
where $f_{feat}$ is the normal encoder layer and $f_{fill}$ is the dedicated residual block for filling the holes. 
\newline To make these residual blocks most efficient in their job, we propose the use adaptive convolutional layers as the building block. It is common to use dilated convolutions or convolutions with large kernels in inpainting networks to extract information from a large neighborhood while filling the holes. In an adaptive convolution layer, each pixel can decide where to look in the neighborhood and how much importance to give to different regions. Utility of such layer has been already established in motion deblurring \cite{purohit2020region}, semantic segmentation \cite{su2019pixel}, etc. We show that such an module is highly beneficial for our approach as the hole regions can adaptively look into different neighborhood regions, both far and near, removing the need for a multi-scale structure. 
\par Given input feature map $x \in \mathbb{R}^{C \times H \times W}$, we apply a kernel and offset generation function to generate a spatially varying kernel $V$ with offsets $\Delta$ and perform the convolutional operation as
\begin{equation}
	y_{j,c} = \sum_{k=1}^K V_{j,j_k} W_c[j_k]x[j+j_k+ \Delta j_k]
\end{equation}
where $K$ is the kernel size, $j_k$ defines position of the convolutional kernel of dilation 1, $V_{j,j_k} \in \mathbb{R}^{K^2 \times H \times W}$ is the pixel dependent kernel generated, $W_c \in \mathbb{R}^{C \times C \times K \times K}$ is the fixed weight and $\Delta j_k$ are the learnable offsets. Note that the kernels ($V$) and offsets ($Delta$) vary from one pixel to another, but are constant for all the channels promoting efficiency. 
%%%%%%%%%%%%%%%%%%%%%%%%% MAIN TABLE %%%%%%%%%%%%%%%%%%%%%%%%%%%%%%%%%%

\begin{table*}\footnotesize

	\centering
	\begin{tabular}{l|l|cc|cc|cc}
		\hline
		\multicolumn{2}{c|}{Dataset} & \multicolumn{2}{|c|}{Places2} & \multicolumn{2}{|c|}{CelebA} & \multicolumn{2}{|c}{Paris Street View}\\
		\hline
		\multicolumn{2}{c|}{Mask Ratio} & 10\%-20\% & 30\%-40\% & 10\%-20\% & 30\%-40\%  & 10\%-20\% & 30\%-40\%   \\
		\hline
		\multirow{6}{*}{SSIM} 
		&PIC         & 0.932 & 0.786 & 0.965 & 0.881  & 0.930 & 0.785 \\
		&PC         & 0.934 & 0.803  & 0.977 & 0.922  & 0.947 & 0.835 \\
		&GC      & ----- & -----& 0.973 & 0.914  & 0.953 & 0.849 \\
		&EC& 0.933 & 0.802  & 0.975 & 0.915  & 0.950 & 0.849 \\
		&Ours                & \textbf{0.939} & \textbf{0.819}  & \textbf{0.981} & \textbf{0.934} & \textbf{0.819}  & \textbf{0.862} \\
		\hline
		\multirow{6}{*}{PSNR} 
		&PIC         & 27.14 & 21.72  & 30.67 & 24.74  & 29.35 & 23.97 \\
		&PC         & 27.29 & 22.12 & 32.77 & 26.94  & 30.76 & 25.46 \\
		&GC      & ----- & -----  & 32.56 & 26.72 & 31.32 & 25.54 \\
		&EC & 27.17 & 22.18  & 32.48 & 26.62  & 31.19 & 26.04 \\
		&Ours                & \textbf{28.75} & \textbf{23.53}  & \textbf{33.86} & \textbf{27.84}  & \textbf{32.71} & \textbf{27.34}\\
		\hline
		
		\hline
	\end{tabular}
	\caption{Numerical comparisons on three datasets.}
%	\vspace{-1\baselineskip}
% 	\vspace{6pt}
	\label{tab:Quantitative_comparison in datases}
\end{table*}

\section{Experimental Results}
\begin{figure*}
	\centering
	\includegraphics[width = 0.75\textwidth]{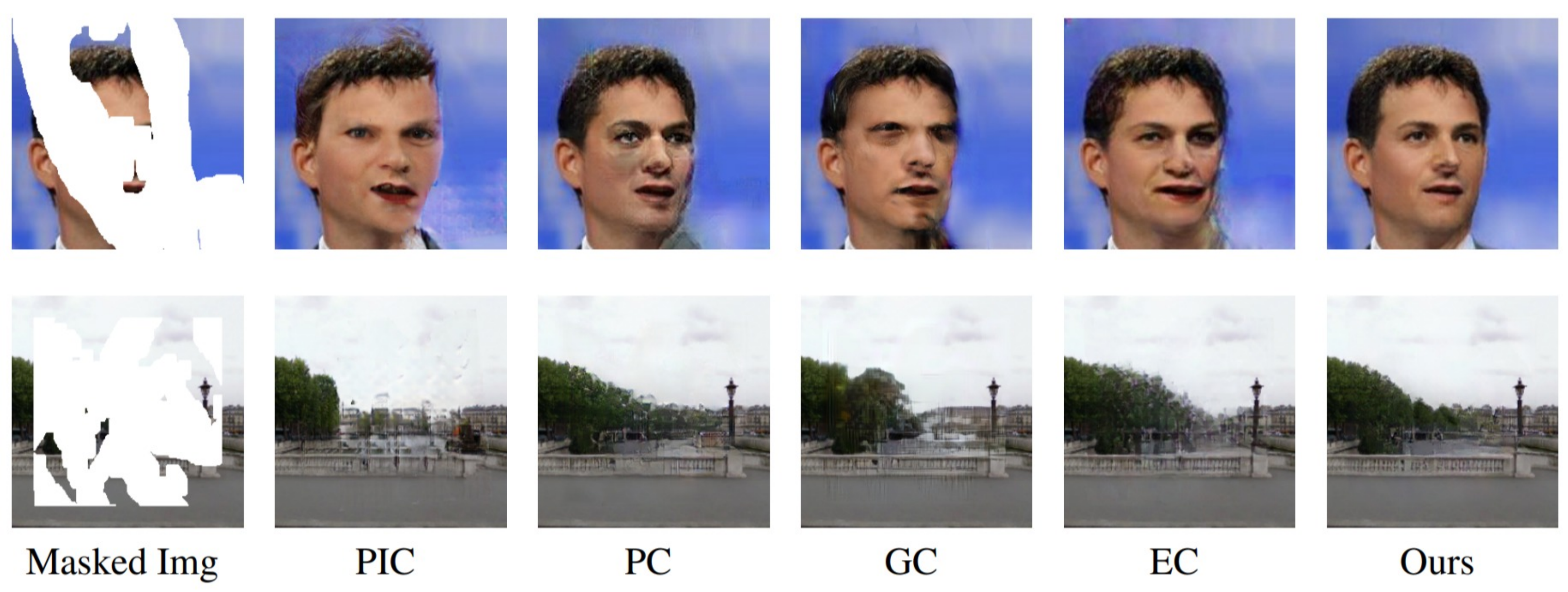}
	\caption{Qualitative results on CelebA and Paris-StreetView.}
	\label{fig:qual}
\end{figure*}

\textbf{Experiment Setup:}
We evaluate our methods on Places2 \cite{zhou2017places}, CelebA \cite{liu2015faceattributes} and Paris StreetView \cite{doersch2012makes} datasets. We use irregular masks from \cite{liu2018image}. The irregular mask dataset contains 12000 irregular masks and the masked area in each mask occupies 0-60\% of the total image size. We train our model with batch size 6 using the Adam optimizer. The approximate number of iterations for CelebA and PSV are 400,000 where as for Places2 it is 2,500,000. We use a standard learning rate of $1e^{-4}$. All experiments are conducted using Pytorch on an Ubuntu 16 system, i7 3.40GHz CPU and an NVIDIA RTX2080Ti GPU.
We compare our approach with several state-of-the-art methods. These models are trained until convergence with the same experiment settings as ours. These models are: PIC (\cite{zheng2019pluralistic}), PC (\cite{liu2018image}), GC (\cite{yu2019free}) and EC (\cite{nazeri2019edgeconnect}).
\newline \textbf{Quantitative Comparisons}: We test all models on the official validation split of Places2, CelebA, and Paris StreetView datasets. We compare our model quantitatively in terms of peak signal-to-noise ratio (PSNR) and structural similarity index (SSIM. Table \ref{tab:Quantitative_comparison in datases} lists the results with different ratios of irregular masks for the three datasets. As shown in Table \ref{tab:Quantitative_comparison in datases}, our method produces excellent results and comfortably surpasses all the comparing models on SSIM, PSNR.
\newline\textbf{Qualitative Comparisons}:
Fig \ref{fig:qual} compare our method with four state-of-the-art approaches on the Places2, CelebA, and Paris StreetView datasets, respectively. Our inpainting results have significantly fewer noticeable inconsistencies in most cases, especially for large holes. Compared to the other methods, our model outperforms the-state-of-the-art with more consistent colors and structures. 

\section{Conclusion}
This work presented a way of improving inpainting network's performance by using knowledge distillation techniques. We investigated an entangled cross and self distillation techniques. Extensive comparisons, ablation studies demonstrates the superiority of the approach. Refined and complete version of this work appeared in ICCV 2021.

\bibliographystyle{unsrtnat}
\bibliography{references}  %%% Uncomment this line and comment out the ``thebibliography'' section below to use the external .bib file (using bibtex) .

\end{document}